\documentclass[11pt]{article}
\usepackage[margin=1in]{geometry}
\usepackage[margin=0.3in]{caption}
\usepackage[utf8]{inputenc}
\usepackage[T1]{fontenc}    
\usepackage{hyperref}       
\usepackage{url}            
\usepackage{booktabs} 
\usepackage{amsfonts} 
\usepackage{nicefrac}   
\usepackage{microtype}
\usepackage{amssymb,amsmath,amsthm,amsfonts}
\usepackage{amsfonts}
\usepackage{graphicx}
\usepackage{color-edits}
\usepackage{algorithm}
\usepackage[noend]{algpseudocode}
\usepackage{caption}
\usepackage{xspace}
\usepackage{esvect}
\usepackage{bbm}
\usepackage{etoolbox}
\newtoggle{submit}
\togglefalse{submit}
\usepackage[numbers, sort,compress]{natbib}

\graphicspath{ {Figures/} }

\newcommand{\reals}{\mathbb{R}}

\newcommand{\vw}{\mathbf{w}}

\newcommand{\vx}{\mathbf{x}}

\newcommand{\R}{\mathbb{R}}

\newcommand{\X}{\mathcal{X}}
\newcommand{\Y}{\mathcal{Y}}

\newcommand{\cP}{\mathcal{P}}

\newtheorem{remark}{Remark}

\newcommand{\jm}[1]{\iftoggle{submit}{}{{\color{red}[Jamie: {#1}]}}}

\newcommand{\sj}[1]{\iftoggle{submit}{}{{\color{blue}[Shahin: {#1}]}}}

\newcommand{\pof}{\textrm{PoF}\xspace}

\newcommand{\x}[1]{\ensuremath{{\bf x}}_{#1}}
\renewcommand{\xi}{\x{i}}
\newcommand{\xj}{\x{j}}
\newcommand{\y}[1]{\ensuremath{y}_{#1}}
\newcommand{\yi}{\y{i}}
\newcommand{\yj}{\y{j}}
\newcommand{\n}[1]{\ensuremath{n}_{#1}}

\newcommand{\mse}{\ensuremath{\textrm{MSE}\xspace}}

\newcommand{\ind}{individual fairness\xspace}
\newcommand{\grp}{group fairness\xspace}
\newcommand{\hybrid}{hybrid fairness\xspace}

\newcommand*{\affaddr}[1]{#1} 
\newcommand*{\affmark}[1][*]{\textsuperscript{#1}}

\begin{document}
\title{\Large\bfseries A Convex Framework for Fair Regression}
\author{
Richard Berk\affmark[1,2], Hoda Heidari\affmark[3], Shahin Jabbari\affmark[3], Matthew Joseph\affmark[3], Michael Kearns\affmark[3], \\
Jamie Morgenstern\affmark[3], Seth Neel\affmark[1], and Aaron Roth\affmark[3]\\
\affaddr{\affmark[1]Department of Statistics}\\
\affaddr{\affmark[2]Department of Criminology}\\
\affaddr{\affmark[3]Department of Computer and Information Science}\\
\affaddr{University of Pennsylvania}\\
}

\maketitle

\begin{abstract}
  We introduce a flexible family of fairness regularizers for (linear
  and logistic) regression problems. These regularizers all enjoy
  convexity, permitting fast optimization, and they span the range
  from notions of group fairness to strong individual
  fairness. By varying the weight on the fairness regularizer, we can
  compute the efficient frontier of the accuracy-fairness trade-off on
  any given dataset, and we measure the severity of this trade-off via
  a numerical quantity we call the Price of Fairness (PoF). The
  centerpiece of our results is an extensive comparative study of the
  PoF across six different datasets in which fairness is a primary
  consideration.
\end{abstract}

\section{Introduction}
\label{sec:intro}

The widespread use of machine learning to make consequential decisions
about individual citizens (including in domains such as credit,
employment, education and criminal sentencing~\cite{hiring, lending,
  policing, sentencing}) has been accompanied by increased reports of
instances in which the algorithms and models employed can be unfair or
discriminatory in a variety of ways~\cite{propublica,Sweeney13}.  As a
result, research on fairness in machine learning and statistics has
seen rapid growth in recent years ~\cite{PedreshiRT08, KamiranCP10,
  CaldersV10, KamiranC11, LuongRT11, KamiranKZ12, KamishimaAAS12,
  HajianD13, FeldmanFMSV15,FishKL16,AdlerFFRSSV16, HardtPS16,
  Chouldechova17, CaldersKKAZ13,DworkHPRZ12}, and several mathematical
formulations have been proposed as metrics of (un)fairness for a
number of different learning frameworks.  While much of the attention
to date has focused on (binary) classification settings, where
standard fairness notions include equal false positive or negative
rates across different populations, less attention has been paid to
fairness in (linear and logistic) regression settings, where the
target and/or predicted values are continuous, and the same value may
not occur even twice in the training data.

In this work, we introduce a rich family of fairness metrics for
regression models that take the form of a fairness regularizer and
apply them to the standard loss functions for linear and logistic
regression.  Since these loss functions and our fairness regularizer
are convex, the combined objective functions obtained from our
framework are also convex, and thus permit efficient
optimization. Furthermore, our family of fairness metrics covers the
spectrum from the type of {\em group\/} fairness that is common in
classification formulations (where e.g. false arrests in one racial
group can be ``compensated'' for by false arrests in another racial
group) to much stronger notions of {\em individual\/} fairness (where
such cancellations are forbidden, and every injustice is charged to
the model). Intermediate fairness notions are also covered. Our
framework also permits one to either forbid the use of a ``protected''
variable (such as race), by demanding that a single model be learned
across all groups, or to build different group-dependent models.

Most importantly, by varying the weight on the fairness regularizer,
our framework permits us to compute the entire ``Pareto curve'' or
efficient frontier of the trade-off between predictive accuracy and
fairness. Such curves are especially important to examine and
understand in a domain-specific manner: since demanding fairness of
models will always come at a cost of reduced predictive
accuracy~\cite{Corbett-DaviesP17, FishKL16, ZafarVGG17, jelveh2015},
it behooves practitioners working with fairness-sensitive data sets to
understand just how mild or severe this trade-off is in their
particular arena, permitting them to make informed modeling and policy
decisions.

Our central results take the form of an extensive comparative
empirical case study across six distinct datasets in which fairness is
a primary concern. For each of these datasets, we compute and examine
the corresponding fairness-accuracy efficiency frontier. We introduce
an intuitive quantity called the {\em Price of Fairness (PoF)\/},
which numerically quantifies the extent to which increased fairness
degrades accuracy. We compare the PoF across datasets, fairness
notions, and treatments of protected variables.

Our primary contributions are:
\begin{itemize}
\item The introduction of a flexible but convex family of fairness
  regularizers of varying strength that spans the spectrum from group
  to individual fairness.
\item The introduction of a quantitative, data-dependent measure of
  the severity of the accuracy-fairness tradeoff.
\item An extensive empirical comparative study across six
  fairness-sensitive data sets.
\end{itemize}

While our empirical study does reveal some reasonably consistent
findings across datasets (e.g. efficiency curves show broadly similar
shapes; PoF generally higher for individual fairness than group;
somewhat surprisingly, PoF not generally improved much when using
protected variables), perhaps the most important message is a
cautionary one: the detailed trade-off between accuracy and fairness,
and the comparison of different fairness notions, appears to be quite
domain-dependent and lacking prescriptive ``universals''. This is
perhaps consistent with the emerging theoretical literature
demonstrating the lack of a single ``right'' definition of
fairness~\cite{KleinbergMR16,Chouldechova17,FriedlerSV16}, and our
work adds evidence to the view that fairness is a topic demanding
careful domain-specific considerations.

\section{The Regression Setting}
Consider the standard (linear and logit) regression setting: denote
the \emph{explanatory} variables (or \emph{instances}) by
$\vx \in \X = \reals^d$ and the \emph{target} variables (or
\emph{labels}) by $y \in \Y = [-1,1]$.  Note that for both linear and
logit models, the target values are continuous. Let $\cP$ denote the
joint distribution over $\X \times \Y$. Suppose every instance $\vx$
belongs to exactly one of $2$ groups, denoted by $1$ and
$2$.\footnote{The generalization to more than 2 groups is
  straightforward.} This partition of $\X$ into groups (e.g. into
different races or genders) is encoded in a ``sensitive'' feature
$\X_{d+1}$. Let $S = \{(\xi, \yi)\}_{i=1}^n\}$ be a training set of
$n$ samples drawn i.i.d. from $\cP$, separated by groups into $S_1$
and $S_2$. Let $n_1 = |S_1|$ and $n_2=|S_2|$.  ($n = n_1+n_2$.)

This work studies the trade-off between fairness and accuracy for the
class of linear and logit regression models.  Given a pair of
explanatory and target variables $(\vx, y)$, we treat $y$ as the
ground truth description of $\vx$'s merit for the regression task at
hand: two pairs $(\vx, y), (\vx', y')$ with $y\approx y'$ have similar
observed outcomes. We aim to design models which treat two such
instances with similar observed outcomes similarly, a notion we refer
to as \emph{ fairness} with respect to the ground truth. For a given
accuracy loss $\ell$ and fairness loss (or \emph{penalty}) $f$, we
define the $\lambda$-weighted fairness loss of a regressor $\vw$ on a
distribution $\cP$ to be $\ell_{\cP}(\vw) + \lambda f_\cP (\vw).$ For
our sample $S$, we analogously define the $\lambda$-weighted fairness
training loss of $\vw$ as $\ell(\vw, S) + \lambda f(\vw, S).$ For
linear regression, we let $\ell$ be mean-squared error; for logistic
regression, we let $\ell$
be the standard log loss. Finally, we use $\ell_2$ regularization for
both models, so the overall loss is then
$ \ell_{\cP}(\vw) + \lambda f_{\cP}(\vw) + \gamma ||\vw||_2 $.

%
%
%

\subsection{A Convex Family of Fairness Regularizers}
Our formal definitions of fairness all measure how similarly a model
treats two similarly labeled instances, one from group $1$ and one
from group $2$.
In particular, all of our definitions have a term for each
``cross-group'' \emph{pair} of instances/labels, weighted as a
function of $|\yi- \yj|$ and also by $|\vw\cdot \xi - \vw \cdot \xj|$.
For shorthand, we will refer to pairs of instances (one from each
group) as \emph{cross pairs}, and cross pairs with similar labels as
similar cross pairs.
Each of the below fairness definitions differs in precisely which
cross pair disparities can counteract one another. In one extreme
(individual fairness in Equation~\ref{eqn:innersq}), every cross pair
disparity increases the fairness penalty of a model. In the other
(group fairness in Equation~\ref{eqn:outersq}), making higher
predictions for the group $1$ instance of a similar cross pair can be
somewhat counterbalanced by making a higher prediction for the group
$2$ instance of a different similar cross pair. Our notions of
fairness for regression align closely to individual and group fairness
definitions for classification, both common threads in the fairness
literature.

\begin{remark}
  We assume the sensitive feature $\X_{d+1}$ is available to the
  learning procedure in one of two ways.  In the first setting, which
  we call the ``single model'' setting, we assume the algorithm builds a
  single linear model $\vw$ for all of $\X$ (over all but the
  sensitive features), but can measure the empirical fairness loss of
  $\vw$ using the sensitive feature. In the second setting, which we
  call the ``separate models'' setting, we allow the algorithm to build two
  distinct linear models $\vw_1, \vw_2$ for the two groups, $\vw_g$
  based on $S_g$, thus directly observing the sensitive feature when
  building these models.
\end{remark}

We specialize the following fairness penalties for the single model
setting, but can easily extend them to the separate models setting, by
replacing $w$ with $w_g$ when applied to a member of group $g$.

\paragraph{Individual Fairness} The first fairness
penalty\footnote{For simplicity we define our fairness penalties on
  samples rather than on the underlying distribution $\cP$.  The
  analogous definitions with respect to the distribution can be
  derived by replacing sums with expectations.}  we propose is the
following:
\begin{align}
f_1(\vw, S) =\frac{1}{\n{1}\n{2}} \sum_{\substack{(\xi, \yi)\in S_1\\ (\xj,\yj)\in S_2}}
 d(\yi,\yj)\big(\vw\cdot \xi-\vw\cdot \xj\big)^2,\label{eqn:innersq}
\end{align}
for some fixed non-negative function $d$, which we assume is
decreasing in $|\yi - \yj|$ (see Section~\ref{sec:exp} for more
details).  Since $d(\yi, \yj)$ does not depend upon the decision
variables ($\vw$), one can treat these values as constants in an
optimization procedure for selecting $\vw$.

The penalty $f_1$ corresponds to \emph{\ind}; for every cross pair
$(\vx,y)\in S_1, (\vx', y')\in S_2$, a model $\vw$ is penalized for
how differently it treats $\vx$ and $\vx'$ (weighted by a function of
$|y - y'|$). No cancellation occurs: the penalty for overestimating
several of one group's labels cannot be mitigated by overestimating
several of the other group's labels.

\paragraph{Group Fairness}
The second fairness penalty we propose is the following:
\begin{align}
f_2(\vw,S) =
\left( \frac{1}{\n{1}\n{2}} \sum_{\substack{(\xi,\yi)\in S_1\\ (\xj,\yj)\in S_2}} 
 d(\yi,\yj) \big(\vw\cdot \xi-\vw\cdot \xj\big)\right)^2.\label{eqn:outersq}
\end{align}

The penalty $f_2$ corresponds to \emph{\grp}: on average, the two
groups' instances should have similar labels (weighted by the nearness
of the labels of the instances). Unlike $f_1$, the penalty $f_2$
allows for \emph{compensation}:
%
informally, if the model over-values some instances of group $1$
relative to group $2$ in similar cross pairs, it can compensate on
other similar cross-pairs by over-valuing those instances from group
$2$ relative to group $1$.

In both of the above formulations, for any cross pair
$(\xi,\yi) \in S_1$ and $(\xj, \yj) \in S_2$, any regressor $\vw$ will
have penalty that increases as
$\left|\vw \cdot \xi-\vw \cdot \xj\right|$ increases, weighted by
$d(\yi, \yj)$.  If the cross pair is similar ($\yi$ is close to $\yj$
and $d(\yi, \yj)$ is large), a regressor which makes very different
predictions for $\xi$ and $\xj$ will incur large loss. If the cross
pair is less similar ($\yi$ is far from $\yj$ and $d(\yi,\yj)$ is
smaller), there is less penalty for having a regressor for which
$\left|\vw \cdot \xi-\vw \cdot \xj\right|$ is large.

\paragraph{Hybrid notions of fairness}
Note that group and individual fairness correspond to two extremes: in
one extreme the fairness penalty considers each cross pair separately
and in the other one the fairness penalty considers all the cross
pairs together.  Mathematically one could define different notions of
fairness by grouping the cross pairs in different manners or even
restrict the fairness penalty only on a subset of cross pairs (called
\emph{bucketing}).  In particular, for binary labeled data (where
$\Y = \{-1,1\}$) one natural choice is to group the cross pairs based
on their labels.  This would result in the following definition of
fairness which we call~\hybrid:
\begin{align}
  f_3(\vw,S) &=
  \left(\sum_{\substack{(\xi,\yi) \in S_{1}\\ (\xj,\yj)\in S_{2}\\ \yi=\yj=1}}
  \frac{d(\yi,\yj) \big(\vw\cdot \xi-\vw\cdot \xj\big)}{n_{1,1} n_{2,1}}\right)^2 + 
  \left( \sum_{\substack{(\xi,\yi) \in S_{1}\\ (\xj,\yj)\in S_{2} \\ \yi=\yj=-1}}
  \frac{d(\yi,\yj) \big(\vw\cdot \xi-\vw\cdot \xj\big)}{n_{1,-1} n_{2,-1}}\right)^2,\label{eqn:betweensq}
\end{align}
where $n_{g,t}$ denotes the size of group $g\in\{1, 2\}$ with label
$t\in\{-1,1\}$ in the sample.  Intuitively, \hybrid~requires both
positive and both negatively labeled cross pairs to be treated
similarly in average over the two groups. Some compensation might
occur, but only amongst instances with the same label: over-valuing
positive instances from group $1$ can mitigate over-valuing positive
instances from group $2$, but does not mitigate under-valuing negative
instances from group $1$. These two terms can be weighted differently
for applications where the treatment of positive (or negative)
instances are more important.  See
Sections~\ref{sec:related}~and~\ref{sec:exp} for more details.


\subsection{Discussion of Our Notions of Fairness}

We now discuss several salient features of our fairness notions.

\paragraph{Why are these fairness notion different from accuracy?}
All of our fairness penalties are small for any perfect regressor (any
$\vw$ such that $\vw \cdot \vx = y$ for all $(\vx,y)\sim\cP$): for a
similar cross pair, $\yi \approx \yj$ and also
$\vw \cdot \xi \approx \vw \cdot \xj$ for a perfect regressor $\vw$.
Our fairness regularizers might then be interpreted as an unusual
proxy for standard accuracy rather than as fairness notions.  However,
perfect (linear or otherwise) regressors almost never exist in
practice; and between two models with similar accuracy, these
definitions bias a learning procedure towards those which have similar
treatment of similarly labeled instances from different groups.

\paragraph{What minimizes these penalties?}
We note that any constant regressor (any $\vw$ such that
$\vw \cdot \vx = c $ for all $\vx\in\X$ and some $c\in \R$) exactly
minimizes all of our fairness regularizers.  As we seen empirically,
this implies that as the fairness regularization factor $\lambda$
increases, we transition from an unfair model with minimum accuracy
loss to a constant (and therefore perfectly fair but trivial) model,
whose accuracy is the best any constant model can achieve.

\section{Related Work}
\label{sec:related}

Recent work has shown that different fairness notions are often
mutually
exclusive~\cite{KleinbergMR16,Chouldechova17,FriedlerSV16}. Unsurprisingly
then, different fairness notions have corresponded to different
algorithms and optimization frameworks.  Previously introduced
fairness notions have generally split along several axis:
classification vs. regression and individual vs. group fairness and
disparate treatment.  Most of previous work has focused on
classification, despite the ubiquity of regression in real world
applications with fairness concerns.

In classification, one line of work aims to achieve the group fairness
notion known as \emph{statistical parity}, i.e. to avoid disparate
impact (see e.g. \cite{PedreshiRT08, KamiranCP10, CaldersV10,
  KamiranC11, LuongRT11, KamiranKZ12, KamishimaAAS12, HajianD13,
  FeldmanFMSV15,FishKL16,AdlerFFRSSV16}).  Statistical parity requires
a predictor to predict each label at similar rates across different
groups.  This definition can be at odds with accuracy especially when
the two groups are inherently different.  \citet{HardtPS16} introduced
a new notion of group fairness called \emph{equality of odds},
partially to alleviate this friction, and partially arguing that
equality of odds more accurately captured what it would mean for a
classifier to be equally ``good'' for two groups. Equality of odds has
a very intuitive interpretation for classification: it requires
similarity of misclassification rates across groups (rather than
forcing the marginal classification rates in the two groups to
coincide). Optimizing for accuracy subject to an equality of odds
constraint was recently shown to be NP-hard \cite{WoodworthGOS17};
work following this result presented efficient heuristics for the
problem \cite{ZafarVGG17, WoodworthGOS17}. We also study
fairness definitions which we can implement efficiently but our interest is in studying
the trade-off between fairness and accuracy and the relationship between different notions
of fairness.

Although equality of odds is also defined for regression, it is very
difficult to determine empirically whether a regressor's output is
conditionally independent of the protected attribute (conditioned on
the true label), as each true label may be seen only once.

\citet{CaldersKKAZ13} introduced the study of statistical parity's
analog in regression settings, (called \emph{equal means} and
\emph{balanced residuals}). More recently \citet{JohnsonFS16} also
studied fairness for regression problems and formalized several
notions for \emph{impartial estimates} based on the causal
relationship between \emph{sensitive attributes}, \emph{legitimate
  attributes}, \emph{suspect attributes} and the label.  Both groups
consider group fairness; our group fairness notion differs from these
as we incorporate the similarity of pairs (through the function $d$)
in our definition though the specific choice of
$d(y,y')=c$ for some $c\in\R$ and all $y$ and $y'$ would recover equal means.

To achieve any of these fairness notions, one needs to decide whether
or not to allow for \emph{disparate treatment} (allowing for different
treatment, or different models, for different groups)\footnote{Any
  classifier that uses sensitive attributes in its decision making is
  implicitly fitting separate models to the two populations, and while
  this might seem unfair, it has been argued that it is actually
  necessary for fairness (most notably in \citet{DworkHPRZ12}).}  , and
where in the learning process to enforce fairness: preprocessing of
data (e.g.~\cite{KamiranC11}); inprocessing, during the training of a
model as either a constraint or incorporated into the objective
function (e.g.~\cite{FishKL16, KamishimaAAS12,ZafarVGG15}); or
postprocessing, where data is labeled by some black-box model and then
relabeled as a function only of the original labels
(e.g. ~\cite{HardtPS16}).  Our approach in this paper falls into the
in-processing category, by encoding fairness as a regularizer (an
approach previously studied in
e.g.~\cite{KamishimaAAS12,ZafarVGG15,ZafarVGG17}). Our work differs
from previous work in several aspects by primarily focusing on
regression, and that our family of fairness measures draws inspiration
from the idea that \emph{similar} instances should be treated
\emph{similarly}~\cite{DworkHPRZ12, ZemelWSPD13}.

\section{A Comparative Empirical Case Study}
\label{sec:exp}

In this section we describe an empirical case study in which we
apply our regularization framework to six different datasets
in which fairness is a central concern. These datasets include
cases in which the observed labels are real-valued, and cases in
which they are binary-valued. For the real-valued datasets, we apply
standard linear regression with our various fairness regularizers.
For the binary-valued datasets, we apply standard logistic regression,
again along with fairness regularizers. 
For datasets with real-valued targets we
normalized the inputs and outputs to be zero mean and unit
variance, and we set the cross-group fairness weights as
$d(\yi, \yj) = e^{-(\yi-\yj)^2}$;
for datasets with binary targets 
we set
$d(\yi,\yj) = \mathbbm{1}[\yi = \yj]$.

For each dataset $S$, our framework requires that we 
solve optimization problems of the
form $\min_{\vw} \ell(\vw,S) + \lambda f(\vw, S) + \gamma ||\vw||_2$ for
variable values of $\lambda$,
where $\ell(\vw,S)$ is either MSE (linear regression)
or the logistic regression loss.
For each $\lambda$ we picked $\gamma$ as
a function of this $\lambda$ by cross validation (see Appendix~\ref{sec:cross} for more details).
All optimization
problems are solved using the CVX solver in Matlab (for real-valued datasets) or
python (for binary-valued datasets).\footnote{See \url{http://www.cvxr.com}~and~\url{http://www.cvxpy.org} for more details. We set the number
of iterations to be 1000 in our optimization solvers.}
Furthermore, all the results are reported using 10-fold cross
validation (see more details in Appendix~\ref{sec:deets}).

The datasets themselves are summarized in Table~\ref{tab:summary},
where we specify the size and dimensionality of each, along with 
the ``protected'' feature (race or gender) 
that thus defines the subgroups across which we apply our
fairness criteria (see Appendix~\ref{sec:data-description} for more details). 
The datasets vary considerably in the
number of observations, their dimensionality, and the relative
size of the minority subgroup.

The \emph{Adult} dataset~\citep{uci-repo, adult-dataset} from the UC Irvine Repository contains 1994
Census data, and the goal is to predict whether the income of an
individual in the dataset is more than $50$K per year or not. 
The sensitive or protected attribute is gender.\footnote{We only used the data in Adult.data in our experiments.}
The \emph{Communities and
  Crime} dataset \citep{uci-repo} includes features relevant to
per capita violent crime rates in different communities in the United
States, and the goal is to predict this crime
rate; race is the protected variable.
The \emph{COMPAS}
dataset contains data from Broward County, Florida, originally compiled
by ProPublica~\cite{propublica}, in which the goal is to predict
whether a convicted individual would commit a violent crime in the
following two years or not. The protected attribute 
is race, and the data was filtered in a fashion similar to that of~\citet{Corbett-DaviesP17}.
The \emph{Default}
dataset~\citep{default-dataset,uci-repo} contains data from Taiwanese
credit card users, and the goal is to predict whether an individual
will default on payments. The protected attribute is
gender.
The \emph{Law School} dataset\footnote{\url{http://www2.law.ucla.edu/sander/Systemic/Data.htm}} 
consists of the records of law students who went on to take the bar exam. The goal is to 
predict whether a student will pass the exam based on features such as LSAT score and 
undergraduate GPA. The protected attribute is gender. 
The \emph{Sentencing} dataset contains information from a state department
of corrections regarding inmates in 2010. The goal is
to predict the sentence length given by the judge based on factors
such as previous criminal records and the crimes for which the
conviction was obtained. The protected attribute is gender.

\begin{table}[h]
\begin{center}
\begin{tabular}{|l|c|c|c|c|c|c|}
\hline
{\bf Data Set\/} & {\bf Type\/}  & {\bf $n$\/} & {\bf $d$\/} & {\bf Minority $n$\/} & {\bf Protected\/}
   \\ \hline
Adult & logit & 32561 & 14 & 10771 & gender
     \\\hline
Communities and Crime & linear & 1994 & 128 & 227 & race
     \\\hline
COMPAS & logit & 3373 & 19 & 1455 & race
     \\\hline
Default & logit & 30000 & 24 & 11888 & gender
     \\\hline
Law School & logit & 27478 & 36 & 12079 & gender
     \\\hline
Sentencing & linear & 5969 & 17  & 385 & gender
     \\\hline
\end{tabular}
\end{center}
\caption{Summary of datasets. 
Type indicates whether regression is logistic or linear;
$n$ is total number of data points;
$d$ is dimensionality;
Minority $n$ is the number of data points in the smaller population;
Protected indicates which feature is protected or fairness-sensitive.
}
\label{tab:summary}
\end{table}

\subsection{Accuracy-Fairness Efficient Frontiers}
\label{sec:pareto}

\begin{figure}[h!]
\begin{minipage}[b]{0.3\textwidth}
\centering
\includegraphics[width=5cm]{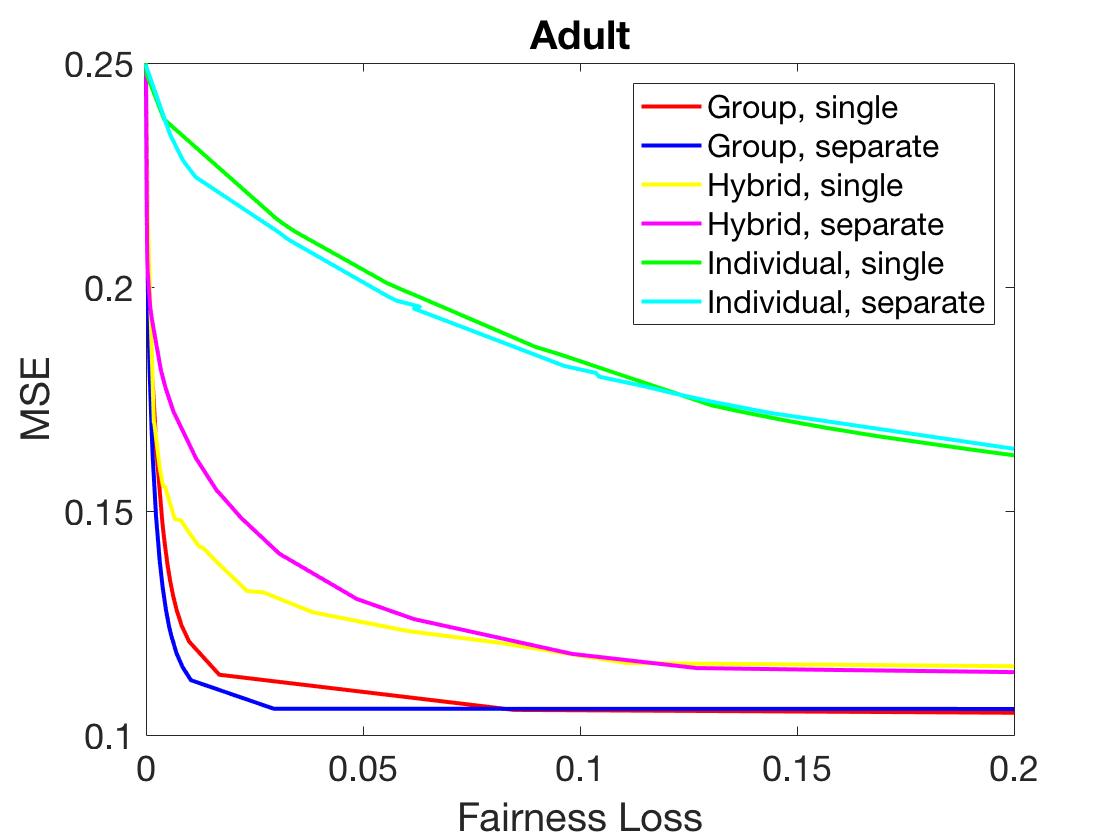}
\end{minipage}
\begin{minipage}[b]{0.3\textwidth}
\includegraphics[width=5cm]{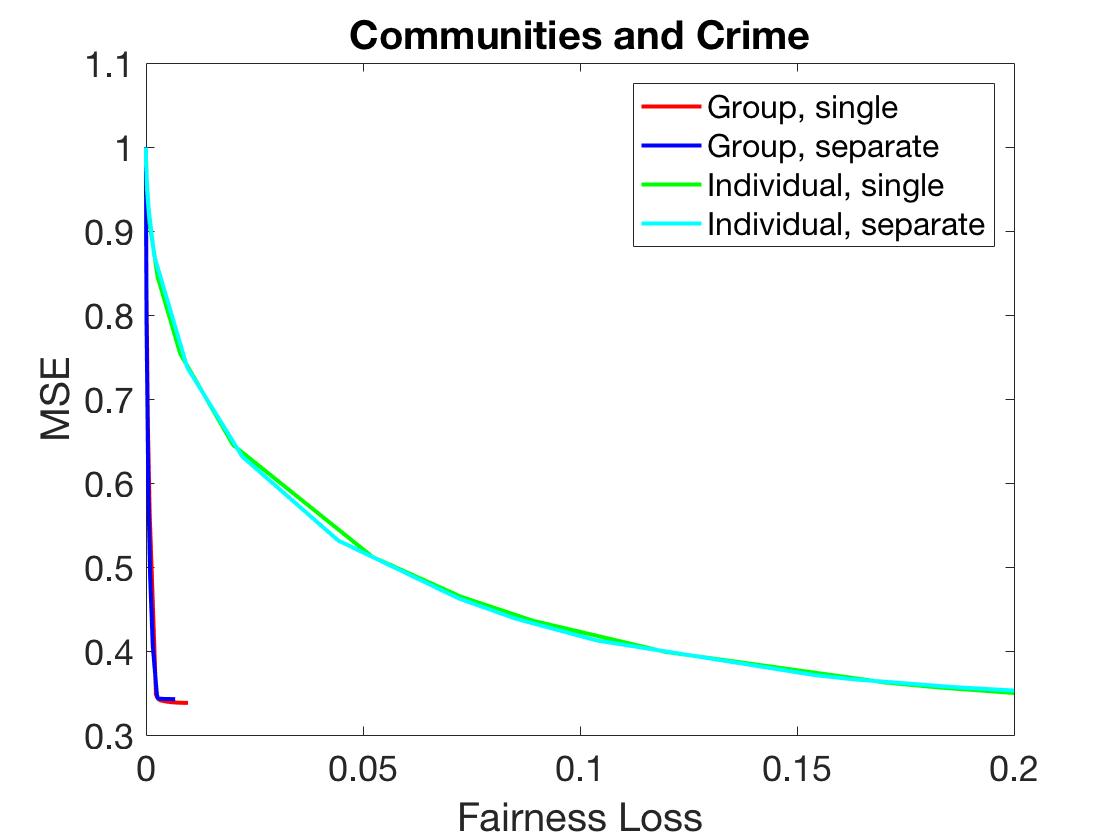}
\end{minipage}
\begin{minipage}[b]{0.3\textwidth}
\includegraphics[width=5cm]{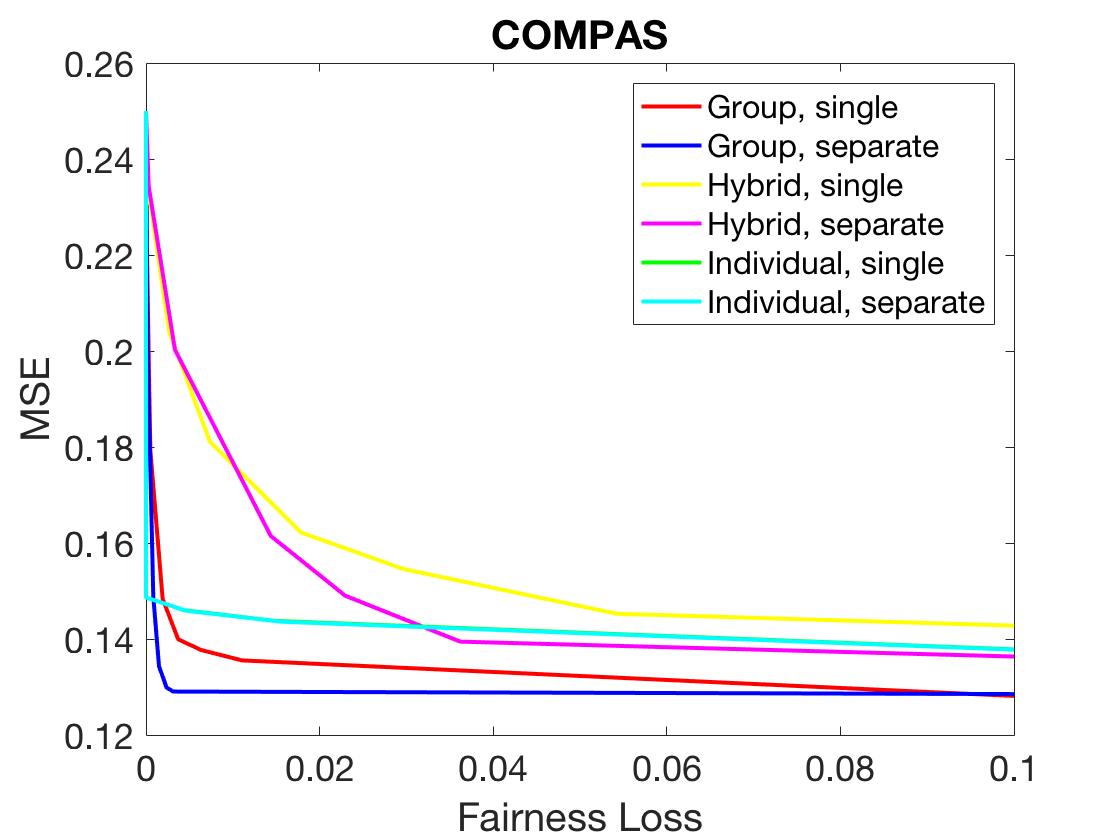}
\end{minipage}\\
\begin{minipage}[b]{0.3\textwidth}
\centering
\includegraphics[width=5cm]{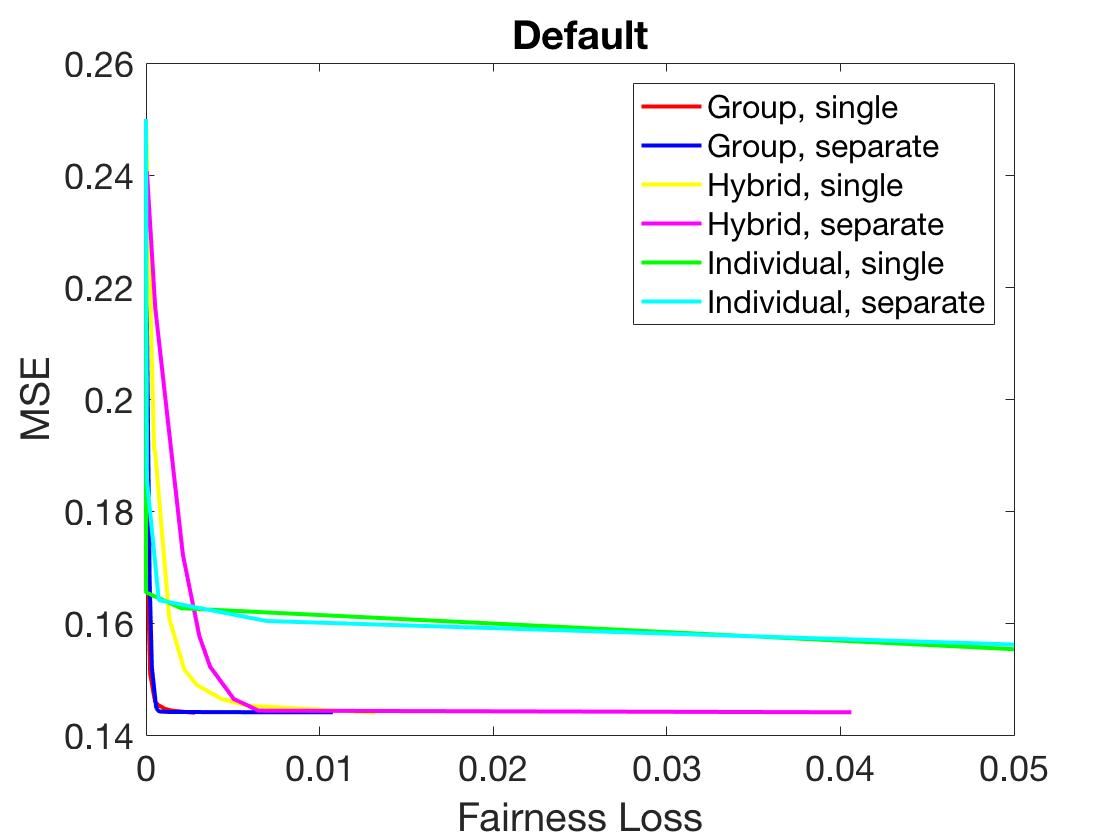}
\end{minipage}
\begin{minipage}[b]{0.3\textwidth}
\includegraphics[width=5cm]{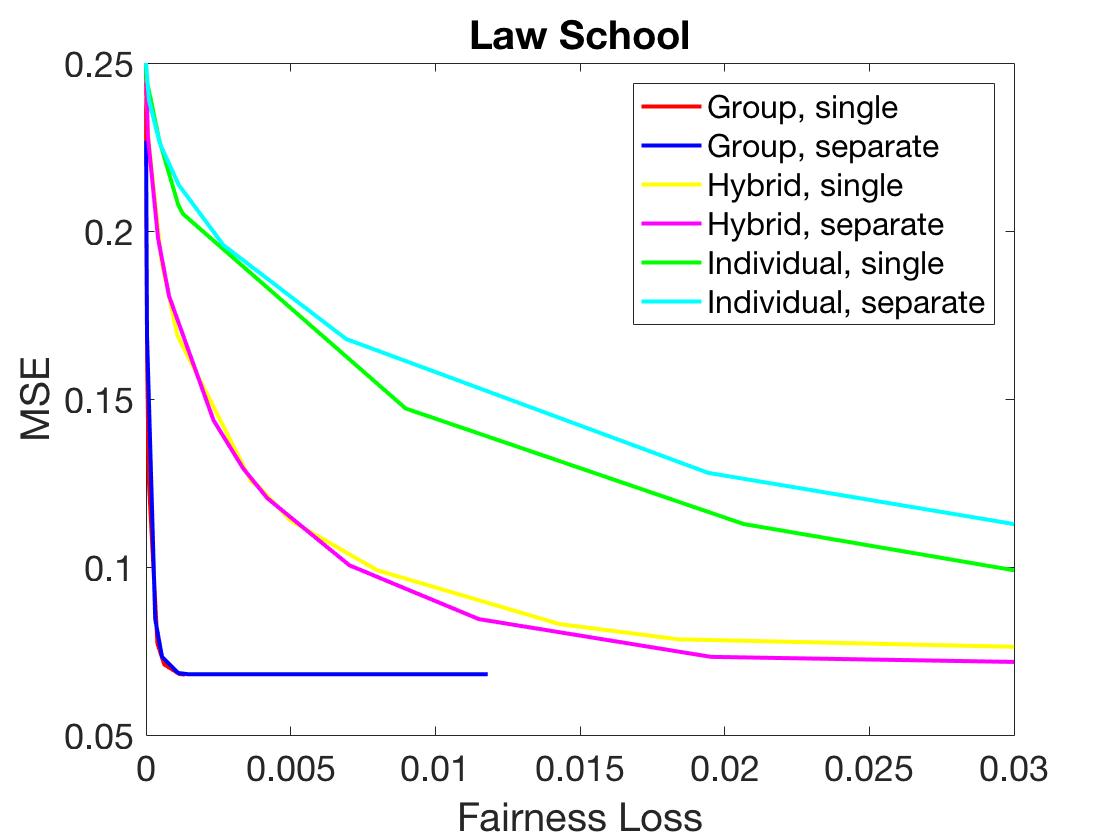}
\end{minipage}
\begin{minipage}[b]{0.3\textwidth}
\includegraphics[width=5cm]{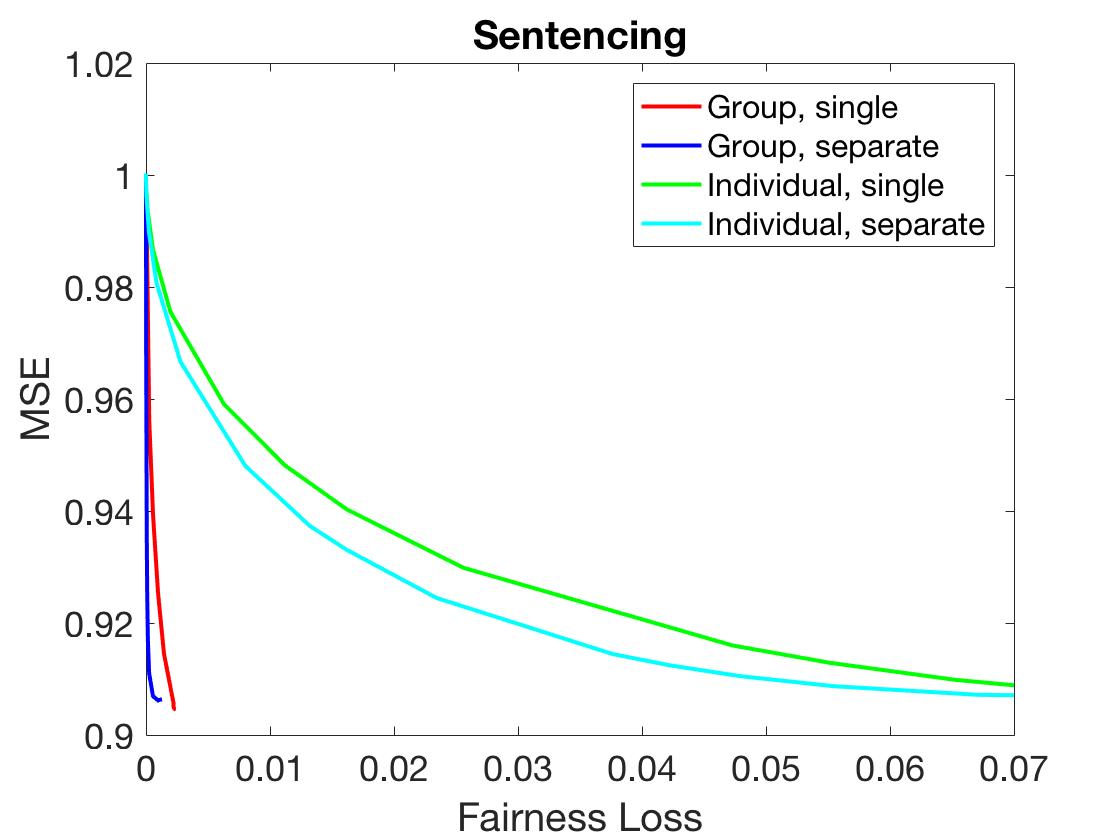}
\end{minipage}
\caption{Efficient frontiers of accuracy vs. fairness for each dataset. For datasets with binary-valued
targets (logistic regression), we consider three fairness notions (group, individual and hybrid), and
for each examine building a single model or separate models for each group, yielding a total of six curves.
For real-valued targets
(linear regression), we consider two fairness notions (group and individual), and again single or
separate models, yielding a total of four curves.
\label{fig:pareto}}
\end{figure}

We begin by examining the efficient frontier of accuracy vs. fairness for
the six datasets.  These curves are shown in Figure~\ref{fig:pareto},
and are obtained by varying the weight $\lambda$ on the fairness
regularizer, and for each value of $\lambda$ finding the model which
minimizes the associated regularized loss function.  For the logistic
regression cases, we extract probabilities from the learned model
$\vw$ as
$\Pr[y_i = 1] = {\exp(\vw \cdot x_i)}/{(1 + \exp(\vw \cdot x_i))}$ and
evaluate these probabilities as predictions for the binary labels
using $\mse$.  \footnote{Note that assessing the $\mse$ of these
  probabilities, interpreted as predictions, is a sensible choice.
  Since squared error is a proper scoring rule, if the labels are
  indeed generated according to a logistic regression model,
  minimizing the squared error of a predictor using mean squared error
  will elicit the true model as its minimizer.}
In all of the datasets, as $\lambda$
increases, the models converge to the best constant predictor,
which minimizes the fairness penalties.

Perhaps the most striking aspect of Figure~\ref{fig:pareto} is the
great diversity of tradeoffs across different datasets and different
fairness regularizers. For instance, if we examine the individual fairness
regularizer, on four of the datasets (Adult, Communities and Crime, Law School
and Sentencing), the curvature is relatively mild and constant --- there is
an approximately fixed rate at which fairness can be traded for accuracy.
In contrast, on COMPAS and Default, fairness loss can be reduced almost for ``free''
until some small threshold value, at which point the accuracy cost increases
dramatically. Similar comments can be made regarding hybrid fairness in the
logistic regression cases.

Individual fairness appears to be strictly more costly than group
fairness for the entire regime between the extremes of $\lambda = 0$
and $\lambda\rightarrow \infty$ for a majority of these datasets (with
the exception of COMPAS and Default datasets). In COMPAS and Default,
for small amounts of unfairness, group unfairness may be more costly
than the same level of individual unfairness.

Perhaps surprisingly, building separate models for each population
barely improves the tradeoff (and in some cases, hurts the
tradeoff for some values of $\lambda$) across almost all datasets and
fairness regularizers. This suggests that the academic discussion
about whether to allow disparate treatment (as explicitly allowed in
e.g. \cite{DworkHPRZ12, JosephKMR16}) -- i.e. whether sensitive
attributes such as race and gender should be ``forbidden'' vs. used in
building more accurate models for each subpopulation, is perhaps less
consequential than expected (at least on these datasets and using
linear or logistic regression models).

Note that in theory \grp is strictly less costly than 
\ind for any particular model $\vw$ (by
Jensen's inequality), and using separate models (one for each group)
should strictly improve the fairness/accuracy trade-off for any of
these notions of fairness. However, both \grp and separate models are more prone to overfitting 
(than \ind and single model), and hence a larger $\ell_2$-regularization parameter $\gamma$ 
tends to be selected in cross validation in these settings (see Appendix~\ref{sec:cross} for more details). 
This is a surprising interaction between the strength of the fairness penalty and the generalization ability 
of the model, and results in group fairness sometimes having a more severe tradeoff with accuracy 
when compared to individual fairness, and separate models having little benefit out of sample, although 
they can appear to have a large effect in-sample (because of the effects of over-fitting).

\newcommand{\C}{\mathcal{C}\xspace}
\subsection{Price of Fairness}
\label{sec:comparison}

The efficient fairness/accuracy frontiers pictured in Figure
\ref{fig:pareto} can be compared across data sets in a qualitative
sense --- e.g. to see that in some datasets, the fairness penalty can
be substantially decreased with little cost to accuracy. However, they
are difficult to compare quantitatively, because the scale of the
fairness loss differs substantially from data set to data set. In this
section, we give a cross-dataset comparison using a measure (which we
call {\em Price of Fairness\/}) which has the effect of normalizing
the fairness loss across data sets to lie on the same scale.

For a given data set and regression type (linear or logistic), let
$\vw^*$ be the optimal model absent any fairness penalty (i.e. the
empirical risk minimizer when the fairness ``regularization'' weight
$\lambda = 0$). This model will suffer some fairness penalty: it
represents the ``maximally unfair'' point on the fairness/accuracy
frontiers from Figure \ref{fig:pareto}. For each dataset, we will fix
a normalization such that this fairness penalty is rescaled to be 1,
and ask for the cost (in terms of the relative increase in mean
squared error) of constraining our predictor to have fairness penalty
$\alpha \leq 1$. Equivalently, this is measuring the relative increase
in $\mse$ that results from constraining a predictor to have fairness
penalty that is no more than \emph{an $\alpha$ fraction of the
  fairness penalty of the unconstrained optimal predictor}.

More formally, let $\vw^* = \arg\min_{\vw} \ell_\cP(\vw)$.
For any value of $\alpha\in[0,1]$ we define the \emph{price of fairness}
(\pof) as follows:
\[
\text{\pof}(\alpha) = \frac{\min_{\vw} \ell_\cP(\vw) \text{ subject to }   f_\cP(\vw) \leq \alpha  f_\cP(\vw^*) }{\ell_\cP(\vw^*)}.
\]

Note that by definition, $\pof(\alpha) \geq 1$, $\pof(1) = 1$, and that $\pof(\alpha)$ increases monotonically as $\alpha$ decreases. Larger values represent more severe costs for imposing fairness constraints that ask that the measure of unfairness be small relative to the unconstrained optimum. It is important to note that because this measure asks for the cost of \emph{relative} improvements over the unconstrained optimum, it can be, for example, that the PoF for one fairness penalty case is larger than for another, even if the \emph{absolute} fairness loss for both the numerator and the denominator is smaller in the second case. With this observation in mind, we can move to the empirical findings.

\begin{figure}[h!]
\begin{minipage}[b]{0.32\textwidth}
\centering
\includegraphics[width=5.8cm]{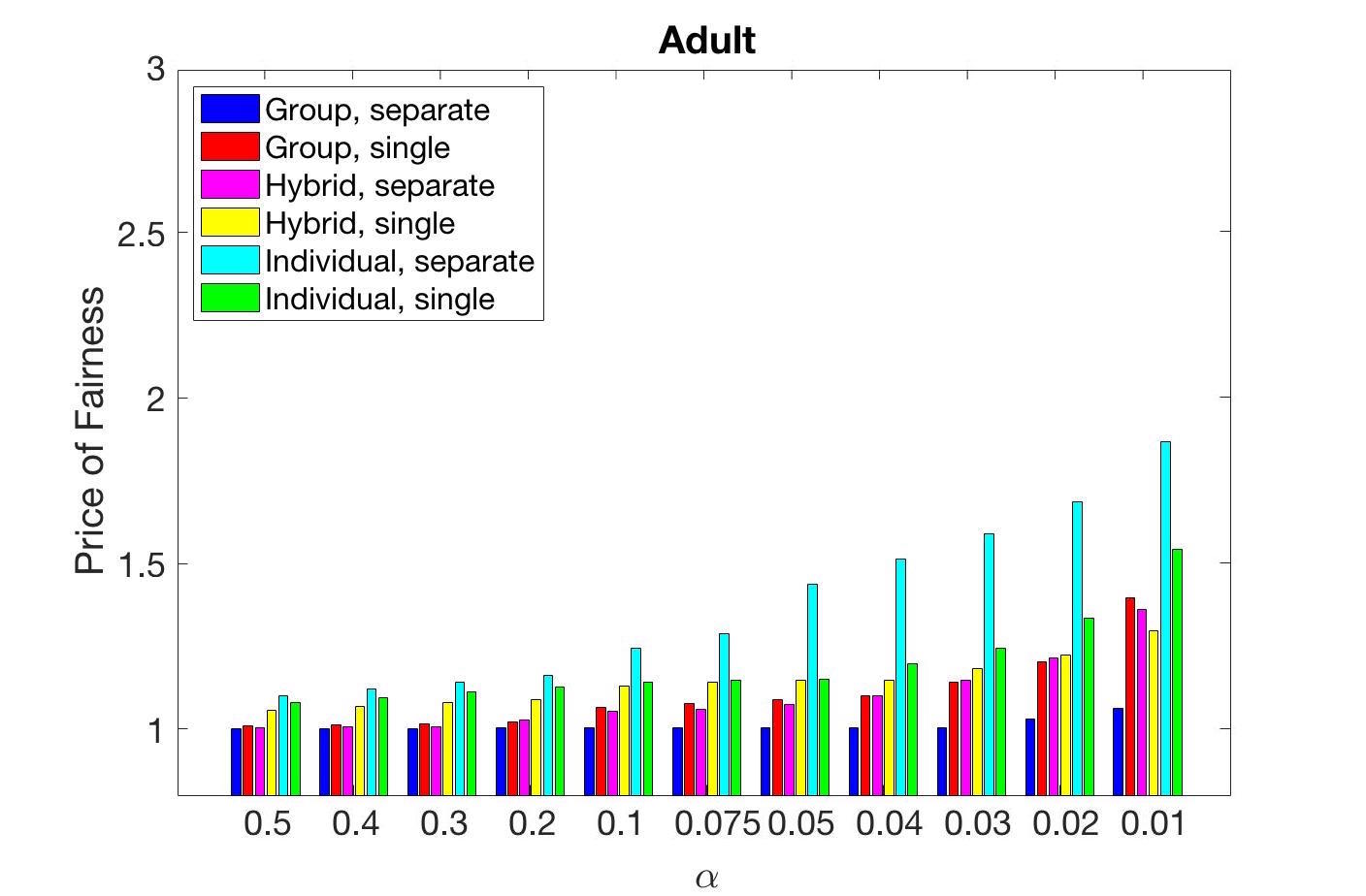}
\end{minipage}
\begin{minipage}[b]{0.32\textwidth}
\includegraphics[width=5.8cm]{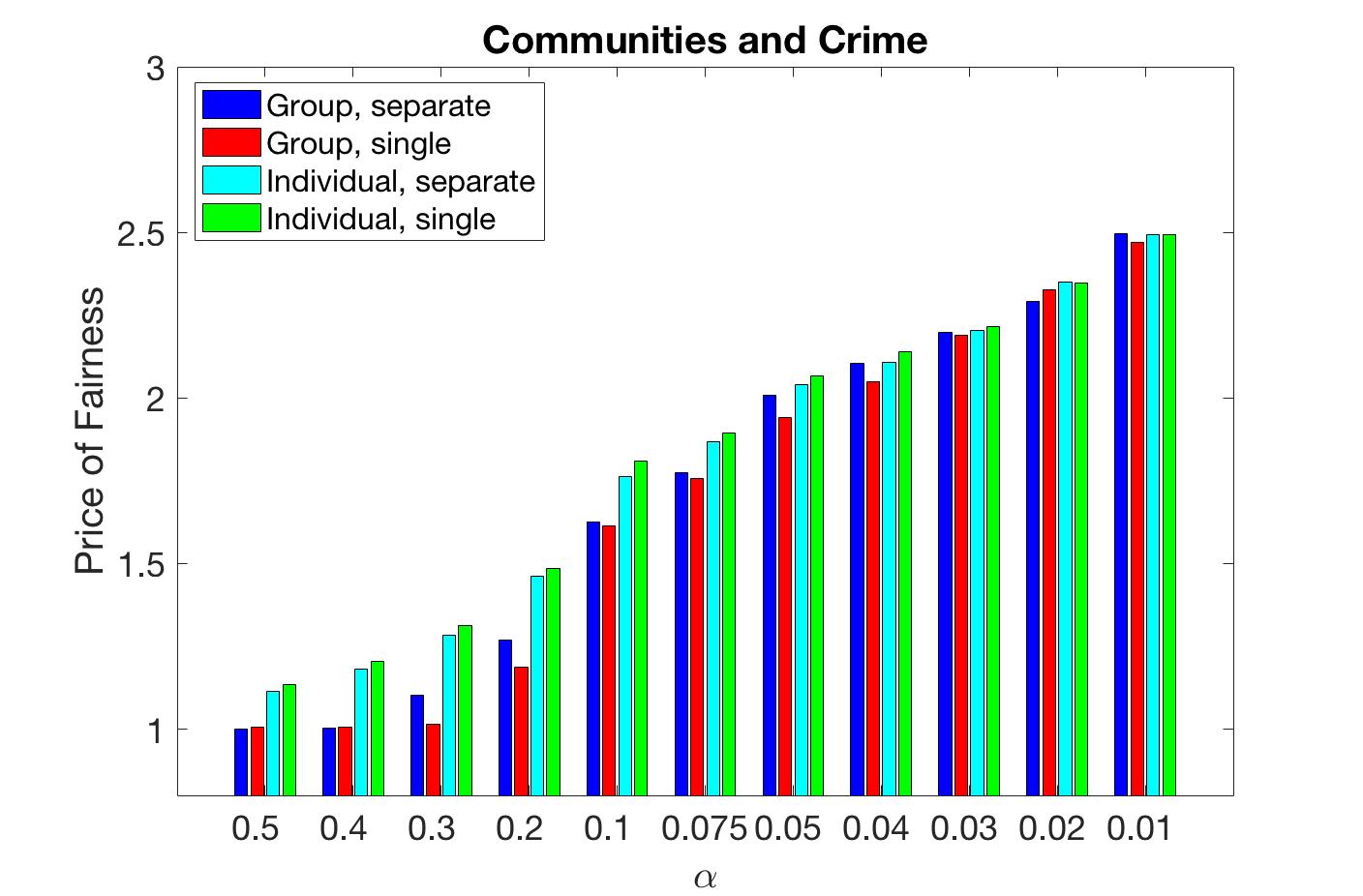}
\end{minipage}
\begin{minipage}[b]{0.32\textwidth}
\includegraphics[width=5.8cm]{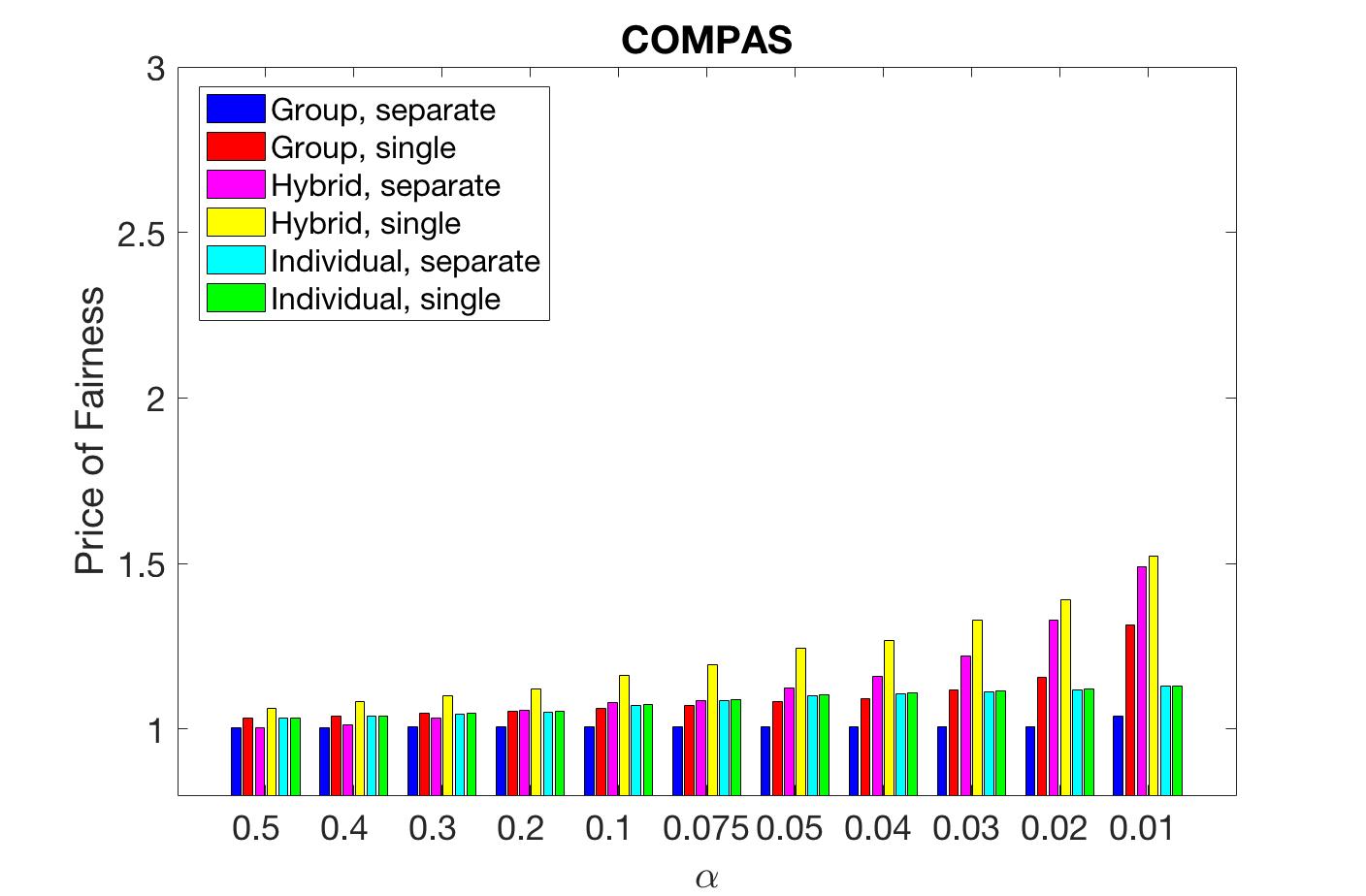}
\end{minipage}\\
\begin{minipage}[b]{0.32\textwidth}
\centering
\includegraphics[width=5.8cm]{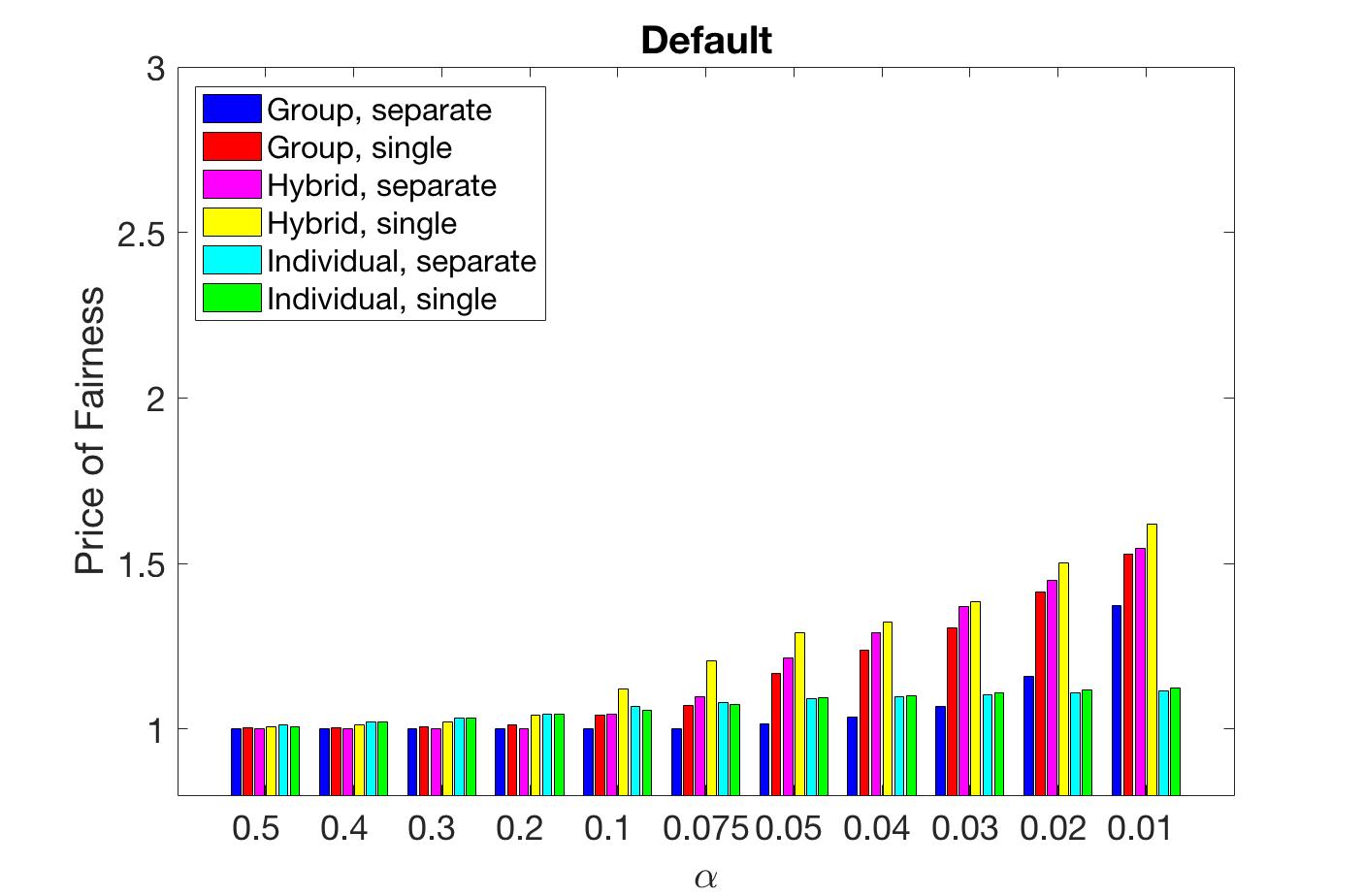}
\end{minipage}
\begin{minipage}[b]{0.32\textwidth}
\includegraphics[width=5.8cm]{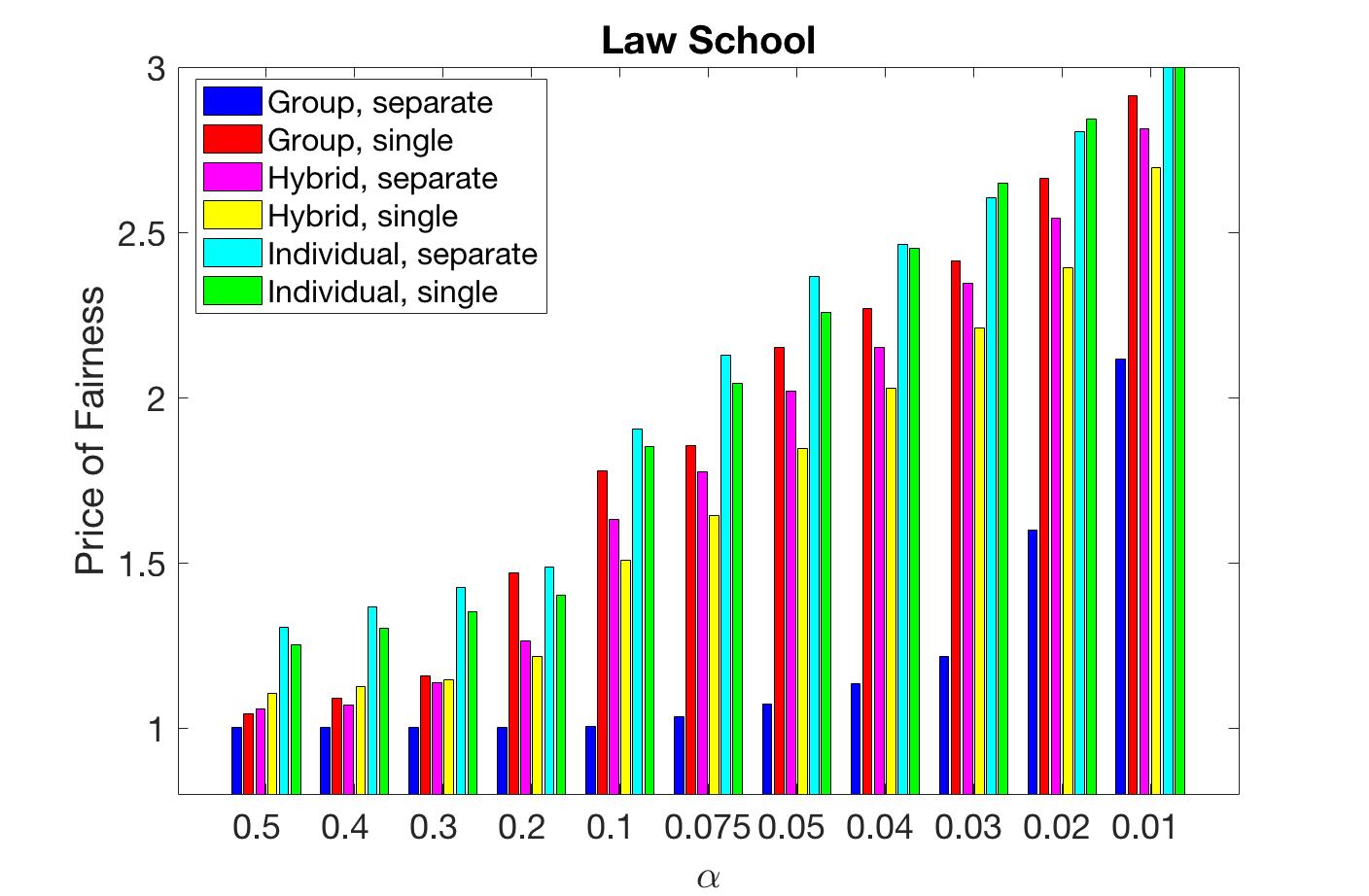}
\end{minipage}
\begin{minipage}[b]{0.32\textwidth}
\includegraphics[width=5.8cm]{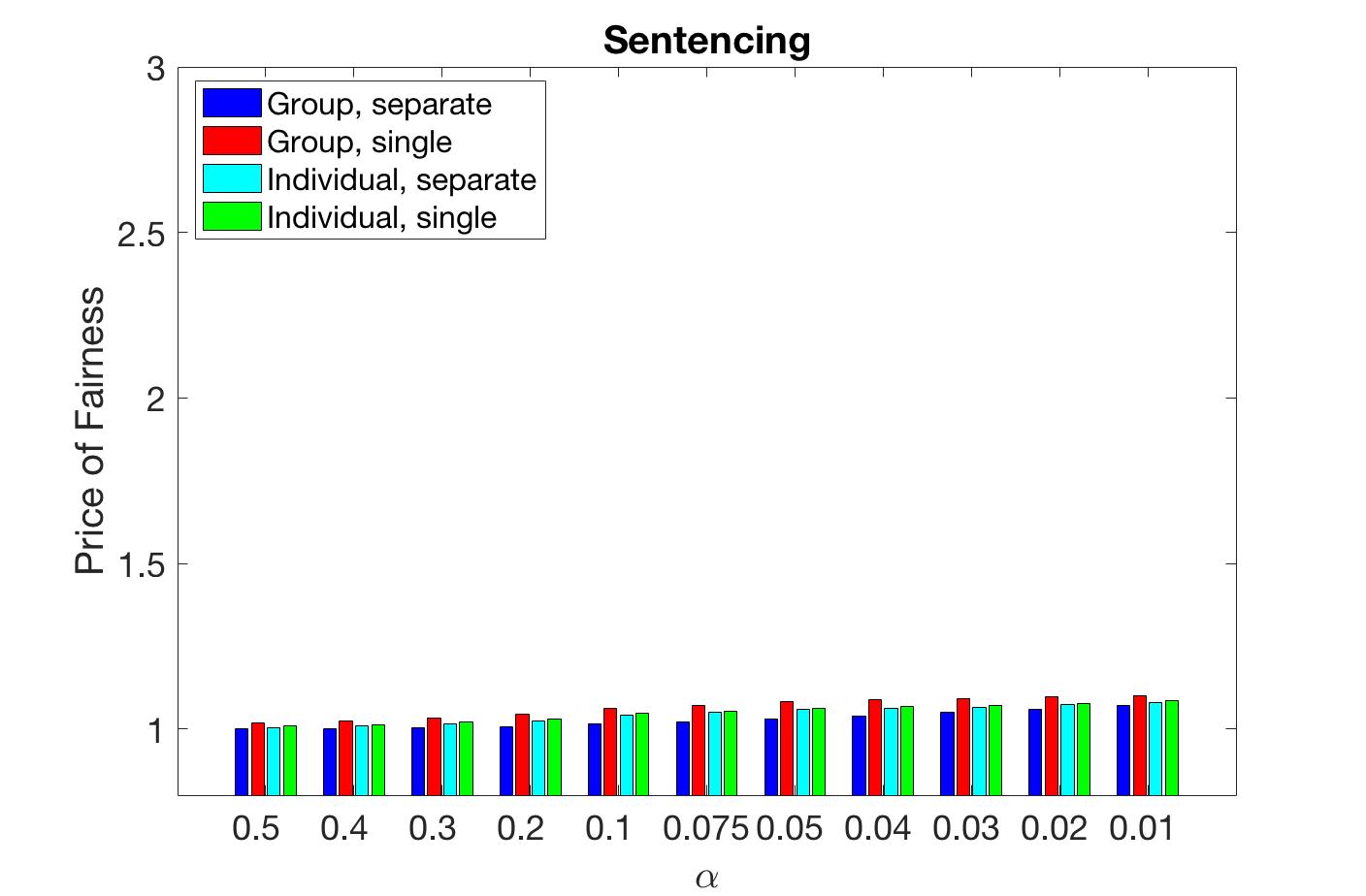}
\end{minipage}
\caption{The ``Price of Fairness'' across data sets, for each type of fairness regularizer, in both the single and separate model case.}   \label{fig:pof}
\end{figure}

Figure~\ref{fig:pof} displays the PoF on each of the $6$ datasets we study, for each fairness regularizer (individual, hybrid, and group), and for the single and separate model case. We first note that even when normalized on a common scale, we continue to see the diversity across datasets that was apparent in Figure~\ref{fig:pareto}. For some datasets (e.g. COMPAS and Sentencing), increasing the fairness constraint by decreasing $\alpha$ has only a mild cost in terms of error. For others (e.g. Communities and Crime, and Law School), the cost increases steadily as we decrease $\alpha$. 

Next, we observe that with this normalization, although the difference between separate and single models remains small on most datasets, on two datasets, differences emerge. In the Law School dataset, restricting to a single model leads to a significantly higher PoF when considering the group fairness metric, compared to allowing separate models. In contrast, on the Adult dataset, restricting to a single model substantially \emph{reduces} the PoF when considering the individual fairness metric. 

Finally, this normalization allows us to observe variation across fairness penalties in the \emph{rate of change} in the PoF as $\alpha$ is decreased. In some datasets (e.g. Communities and Crime, and Sentencing), the PoF changes in lock-step across all measures of unfairness. However, for others (e.g. Default), the PoF increases substantially with $\alpha$ when we consider group or hybrid fairness measures, but is much more stable for individual fairness. 

\section{Conclusions}

The use of a complexity regularizer to control overfitting is both
standard and well-understood in machine learning. While the use of
such a regularizer introduces a trade-off --- goodness of fit
vs. model complexity --- it does not introduce a {\em tension\/},
because complexity regularization is always in service of improving
generalization, and is usually not a goal in its own right.

In contrast, in this work we have studied a variety of {\em
  fairness\/} regularizers for regression problems, and applied them
to data sets in which fairness is not subservient to generalization,
but is instead a first-order consideration.  Our empirical study has
demonstrated that the choice of fairness regularizer (group,
individual, hybrid, or other) and the particular data set can have
qualitative effects on the trade-off between accuracy and
fairness. Combined with recent theoretical
results~\cite{KleinbergMR16,Chouldechova17,FriedlerSV16} that also
highlight the incompatibility of various fairness measures, our
results highlight the care that must be taken by practitioners in
defining the type of fairness they care about for their particular
application, and in determining the appropriate balance between
predictive accuracy and fairness.

\bibliographystyle{plainnat}
\bibliography{bib}

\begin{thebibliography}{35}
\providecommand{\natexlab}[1]{#1}
\providecommand{\url}[1]{\texttt{#1}}
\expandafter\ifx\csname urlstyle\endcsname\relax
  \providecommand{\doi}[1]{doi: #1}\else
  \providecommand{\doi}{doi: \begingroup \urlstyle{rm}\Url}\fi

\bibitem[Adler et~al.(2016)Adler, Falk, Friedler, Rybeck, Scheidegger, Smith,
  and Venkatasubramanian]{AdlerFFRSSV16}
Philip Adler, Casey Falk, Sorelle Friedler, Gabriel Rybeck, Carlos Scheidegger,
  Brandon Smith, and Suresh Venkatasubramanian.
\newblock Auditing black-box models for indirect influence.
\newblock In \emph{Proceedings of the 16th International Conference on Data
  Mining}, pages 1--10, 2016.

\bibitem[Angwin et~al.(2016)Angwin, Larson, Mattu, and Kirchner]{propublica}
Julia Angwin, Jeff Larson, Surya Mattu, and Lauren Kirchner.
\newblock Machine bias.
\newblock \emph{Propublica}, 2016.

\bibitem[Barry-Jester et~al.(2015)Barry-Jester, Casselman, and
  Goldstein]{sentencing}
Anna Barry-Jester, Ben Casselman, and Dana Goldstein.
\newblock The new science of sentencing.
\newblock \emph{The Marshall Project}, August 8 2015.
\newblock URL
  \url{https://www.themarshallproject.org/2015/08/04/the-new-science-of-sentencing}.
\newblock Retrieved 4/28/2016.

\bibitem[Byrnes(2016)]{lending}
Nanette Byrnes.
\newblock Artificial intolerance.
\newblock \emph{MIT Technology Review}, March 28 2016.
\newblock URL
  \url{https://www.technologyreview.com/s/600996/artificial-intolerance/}.
\newblock Retrieved 4/28/2016.

\bibitem[Calders and Verwer(2010)]{CaldersV10}
Toon Calders and Sicco Verwer.
\newblock Three naive {Bayes} approaches for discrimination-free
  classification.
\newblock \emph{Data Mining and Knowledge Discovery}, 21\penalty0 (2):\penalty0
  277--292, 2010.

\bibitem[Calders et~al.(2013)Calders, Karim, Kamiran, Ali, and
  Zhang]{CaldersKKAZ13}
Toon Calders, Asim Karim, Faisal Kamiran, Wasif Ali, and Xiangliang Zhang.
\newblock Controlling attribute effect in linear regression.
\newblock In \emph{Proceedings of the 13th International Conference on Data
  Mining}, pages 71--80, 2013.

\bibitem[Chouldechova(2017)]{Chouldechova17}
Alexandra Chouldechova.
\newblock Fair prediction with disparate impact: A study of bias in recidivism
  prediction instruments.
\newblock \emph{coRR}, abs/1703.00056, 2017.

\bibitem[Corbett{-}Davies et~al.(2017)Corbett{-}Davies, Pierson, Feller, Goel,
  and Huq]{Corbett-DaviesP17}
Sam Corbett{-}Davies, Emma Pierson, Avi Feller, Sharad Goel, and Aziz Huq.
\newblock Algorithmic decision making and the cost of fairness.
\newblock \emph{CoRR}, abs/1701.08230, 2017.

\bibitem[Dwork et~al.(2012)Dwork, Hardt, Pitassi, Reingold, and
  Zemel]{DworkHPRZ12}
Cynthia Dwork, Moritz Hardt, Toniann Pitassi, Omer Reingold, and Richard Zemel.
\newblock Fairness through awareness.
\newblock In \emph{Proceedings of the 3rd Conference on Innovations in
  Theoretical Computer Science}, pages 214--226, 2012.

\bibitem[Feldman et~al.(2015)Feldman, Friedler, Moeller, Scheidegger, and
  Venkatasubramanian]{FeldmanFMSV15}
Michael Feldman, Sorelle Friedler, John Moeller, Carlos Scheidegger, and Suresh
  Venkatasubramanian.
\newblock Certifying and removing disparate impact.
\newblock In \emph{Proceedings of the 21th {ACM} {SIGKDD} International
  Conference on Knowledge Discovery and Data Mining}, pages 259--268, 2015.

\bibitem[Fish et~al.(2016)Fish, Kun, and Lelkes]{FishKL16}
Benjamin Fish, Jeremy Kun, and {\'{A}}d{\'{a}}m~D{\'{a}}niel Lelkes.
\newblock A confidence-based approach for balancing fairness and accuracy.
\newblock In \emph{Proceedings of the 16th {SIAM} International Conference on
  Data Mining}, pages 144--152, 2016.

\bibitem[Friedler et~al.(2016)Friedler, Scheidegger, and
  Venkatasubramanian]{FriedlerSV16}
Sorelle Friedler, Carlos Scheidegger, and Suresh Venkatasubramanian.
\newblock On the (im)possibility of fairness.
\newblock \emph{CoRR}, abs/1609.07236, 2016.

\bibitem[Hajian and Domingo{-}Ferrer(2013)]{HajianD13}
Sara Hajian and Josep Domingo{-}Ferrer.
\newblock A methodology for direct and indirect discrimination prevention in
  data mining.
\newblock \emph{{IEEE} Transactions on Knowledge and Data Engineering},
  25\penalty0 (7):\penalty0 1445--1459, 2013.

\bibitem[Hardt et~al.(2016)Hardt, Price, and Srebro]{HardtPS16}
Moritz Hardt, Eric Price, and Nathan Srebro.
\newblock Equality of opportunity in supervised learning.
\newblock In \emph{Proceedings of the 30th Annual Conference on Neural
  Information Processing Systems}, pages 3315--3323, 2016.

\bibitem[Jelveh and Luca(2015)]{jelveh2015}
Zubin Jelveh and Michael Luca.
\newblock Towards diagnosing accuracy loss in discrimination-aware
  classification: An application to predictive policing.
\newblock In \emph{2nd Workshop on Fairness, Accountability, and Transparency
  in Machine Learning}, 2015.

\bibitem[Johnson et~al.(2016)Johnson, Foster, and Stine]{JohnsonFS16}
Kory Johnson, Dean Foster, and Robert Stine.
\newblock Impartial predictive modeling: Ensuring fairness in arbitrary models.
\newblock \emph{CoRR}, abs/1608.00528, 2016.

\bibitem[Joseph et~al.(2016)Joseph, Kearns, Morgenstern, and Roth]{JosephKMR16}
Matthew Joseph, Michael Kearns, Jamie Morgenstern, and Aaron Roth.
\newblock Fairness in learning: classic and contextual bandits.
\newblock In \emph{Proceedings of The 30th Annual Conference on Neural
  Information Processing Systems}, pages 325--333, 2016.

\bibitem[Kamiran and Calders(2011)]{KamiranC11}
Faisal Kamiran and Toon Calders.
\newblock Data preprocessing techniques for classification without
  discrimination.
\newblock \emph{Knowledge and Information Systems}, 33\penalty0 (1):\penalty0
  1--33, 2011.

\bibitem[Kamiran et~al.(2010)Kamiran, Calders, and Pechenizkiy]{KamiranCP10}
Faisal Kamiran, Toon Calders, and Mykola Pechenizkiy.
\newblock Discrimination aware decision tree learning.
\newblock In \emph{Proceedings of the 10th {IEEE} International Conference on
  Data Mining}, pages 869--874, 2010.

\bibitem[Kamiran et~al.(2012)Kamiran, Karim, and Zhang]{KamiranKZ12}
Faisal Kamiran, Asim Karim, and Xiangliang Zhang.
\newblock Decision theory for discrimination-aware classification.
\newblock In \emph{Proceedings of the 12th {IEEE} International Conference on
  Data Mining}, pages 924--929, 2012.

\bibitem[Kamishima et~al.(2012)Kamishima, Akaho, Asoh, and
  Sakuma]{KamishimaAAS12}
Toshihiro Kamishima, Shotaro Akaho, Hideki Asoh, and Jun Sakuma.
\newblock Fairness-aware classifier with prejudice remover regularizer.
\newblock In \emph{Proceedings of the European Conference on Machine Learning
  and Knowledge Discovery in Databases}, pages 35--50, 2012.

\bibitem[Kleinberg et~al.(2017)Kleinberg, Mullainathan, and
  Raghavan]{KleinbergMR16}
Jon Kleinberg, Sendhil Mullainathan, and Manish Raghavan.
\newblock Inherent trade-offs in the fair determination of risk scores.
\newblock In \emph{Proceedings of the 8th Conference on Innovations in
  Theoretical Computer Science}, 2017.

\bibitem[Kohavi(1996)]{adult-dataset}
Ron Kohavi.
\newblock Scaling up the accuracy of naive {Bayes} classifiers: {A}
  decision-tree hybrid.
\newblock In \emph{Proceedings of the 2nd International Conference on Knowledge
  Discovery and Data Mining}, pages 202--207, 1996.

\bibitem[Lichman(2013)]{uci-repo}
Moshe Lichman.
\newblock {UCI} machine learning repository, 2013.
\newblock URL \url{http://archive.ics.uci.edu/ml}.

\bibitem[Luong et~al.(2011)Luong, Ruggieri, and Turini]{LuongRT11}
Binh~Thanh Luong, Salvatore Ruggieri, and Franco Turini.
\newblock k-{NN} as an implementation of situation testing for discrimination
  discovery and prevention.
\newblock In \emph{Proceedings of the 17th ACM SIGKDD international conference
  on Knowledge discovery and data mining}, pages 502--510. ACM, 2011.

\bibitem[Miller(2015)]{hiring}
Clair Miller.
\newblock Can an algorithm hire better than a human?
\newblock \emph{The New York Times}, June 25 2015.
\newblock URL
  \url{http://www.nytimes.com/2015/06/26/upshot/can-an-algorithm-hire-better-than-a-human.html/}.
\newblock Retrieved 4/28/2016.

\bibitem[Pedreshi et~al.(2008)Pedreshi, Ruggieri, and Turini]{PedreshiRT08}
Dino Pedreshi, Salvatore Ruggieri, and Franco Turini.
\newblock Discrimination-aware data mining.
\newblock In \emph{Proceedings of the 14th ACM SIGKDD international conference
  on Knowledge discovery and data mining}, pages 560--568. ACM, 2008.

\bibitem[Redmond and Baveja(2002)]{communities-crime-dataset}
Michael Redmond and Alok Baveja.
\newblock A data-driven software tool for enabling cooperative information
  sharing among police departments.
\newblock \emph{European Journal of Operational Research}, 141\penalty0
  (3):\penalty0 660--678, 2002.

\bibitem[Rudin(2013)]{policing}
Cynthia Rudin.
\newblock Predictive policing using machine learning to detect patterns of
  crime.
\newblock \emph{Wired Magazine}, August 2013.
\newblock URL
  \url{http://www.wired.com/insights/2013/08/predictive-policing-using-machine-learning-to-detect-
  patterns-of-crime/}.
\newblock Retrieved 4/28/2016.

\bibitem[Sweeney(2013)]{Sweeney13}
Latanya Sweeney.
\newblock Discrimination in online ad delivery.
\newblock \emph{Communications of the {ACM}}, 56\penalty0 (5):\penalty0 44--54,
  2013.

\bibitem[Woodworth et~al.(2017)Woodworth, Gunasekar, Ohannessian, and
  Srebro]{WoodworthGOS17}
Blake Woodworth, Suriya Gunasekar, Mesrob Ohannessian, and Nathan Srebro.
\newblock Learning non-discriminatory predictors.
\newblock \emph{CoRR}, abs/1702.06081, 2017.

\bibitem[Yeh and Lien(2009)]{default-dataset}
I-Cheng Yeh and Che-hui Lien.
\newblock The comparisons of data mining techniques for the predictive accuracy
  of probability of default of credit card clients.
\newblock \emph{Expert Systems with Applications}, 36\penalty0 (2):\penalty0
  2473--2480, 2009.

\bibitem[Zafar et~al.(2015)Zafar, Valera, Gomez{-}Rodriguez, and
  Gummadi]{ZafarVGG15}
Muhammad Zafar, Isabel Valera, Manuel Gomez{-}Rodriguez, and Krishna Gummadi.
\newblock Fairness constraints: {A} mechanism for fair classification.
\newblock \emph{CoRR}, abs/1507.05259, 2015.

\bibitem[Zafar et~al.(2017)Zafar, Valera, Gomez{-}Rodriguez, and
  Gummadi]{ZafarVGG17}
Muhammad~Bilal Zafar, Isabel Valera, Manuel Gomez{-}Rodriguez, and Krishna~P.
  Gummadi.
\newblock Fairness beyond disparate treatment {\&} disparate impact: Learning
  classification without disparate mistreatment.
\newblock In \emph{Proceedings of the 26th International Conference on World
  Wide Web}, pages 1171--1180, 2017.

\bibitem[Zemel et~al.(2013)Zemel, Wu, Swersky, Pitassi, and Dwork]{ZemelWSPD13}
Richard Zemel, Yu~Wu, Kevin Swersky, Toni Pitassi, and Cynthia Dwork.
\newblock Learning fair representations.
\newblock In \emph{Proceedings of the 30th International Conference on Machine
  Learning}, pages 325--333, 2013.

\end{thebibliography}

\appendix
\section{Missing Details from the Experiments}
\label{sec:exp-appendix}

\subsection{Cross Validation for Picking $\gamma$}
\label{sec:cross}

In this section we show how we used cross validation in our experiments to find $\gamma$.
For each dataset $S$, our framework requires that we 
solve optimization problems of the
form $\min_{\vw} \ell(\vw,S) + \lambda f(\vw, S) + \gamma ||\vw||_2$ for
variable values of $\lambda$,
where $\ell(\vw,S)$ is either MSE (linear regression)
or the logistic regression loss.
For each $\lambda$ we picked $\gamma$ as
a function of this $\lambda$ as follows:

We divided the data into 10 folds (partitions).
Let $S_i$ denote the data of fold $i$ and
$S_{-i}$ denote all the data except fold $i$. 
Let $\Gamma=\{\gamma_1, \ldots, \gamma_k\}$ be a set of $k$ potential values for $\gamma$
we are selecting from and 
$L$ be a vector of size $k$ initialized to all zero.

\begin{algorithm}[h]
 \begin{algorithmic}
\State \For{$j = 1, \ldots, k$ (repeat over values of $\gamma$ in $\Gamma$)}
\State \For{$i = 1, \ldots, 10$ (10-fold cross-validation loop)}
\State Compute $w_i(\gamma_j) \in \arg\min_{w} \ell(w, S_{-i}) + \lambda f(w, S_{-i}) + \gamma_j\|w\|_2$
\State \Comment{train a model $w_i(\gamma_j)$ on all the data except fold $i$}
\State Set $\text{Loss}(i,j) = \ell(w_i(\gamma_j), S_{i}) + \lambda f(w_i(\gamma_j), S_{i})$
\State \Comment{record the test loss of the computed model $w_i(\gamma_j)$ on fold $i$} 
\State $L(j) = L(j) + \text{Loss}(i,j)$.
\State \Comment{add the test loss to the total loss of $\gamma_j$ so far.}
\EndFor
\EndFor
\end{algorithmic}
\caption{Procedure for Picking $\gamma$}
\label{alg:lambda}
\end{algorithm}

 After the loops are over, let $k^*$ denote the entry of $L$ with the smallest value. Then we pick $\gamma_{k^*}\equiv \gamma(\lambda)$ as the regularizer.
So our regularized objective function for this particular $\lambda$ becomes 
\[
\min_w \ell(w,S) + \lambda f(w,S) + \gamma_{\lambda} \|w\|
\]

We observed in our experiments that $\gamma$ increases as $\lambda$ increases. 
Moreover, for any fixed $\lambda$, the value of $\gamma$ chosen for separate models is usually higher than the value
of $\gamma$ chosen for single model.

%
%
%
%
%
\subsection{Additional Details}
\label{sec:deets}

We used 10-fold cross validation to evaluate the performance (fairness
and accuracy losses) of a trained model on test data. For smaller
datasets (Communities and Crime, COMPAS and Sentencing) we used one
run of 10-fold cross validation. For larger datasets (Adult, Default
and Law School) we first randomly sampled 30-50\% of the dataset
(depending on the dataset), ran the 10-fold cross validation on the
sampled data and then repeated the experiment 3 times each time with a
new random sample. This is because both the running time and the
instability of the CVXPY solver would increase when we used the whole
datasets for the larger datasets.

While the fairness losses~\ref{eqn:innersq}, \ref{eqn:outersq}, and
\ref{eqn:betweensq} are defined using all the $n_1\times n_2$ cross
pairs in the dataset, in our experiments we only used
$2 \times \text{Minority } n$ random cross pairs where
$\text{Minority } n = \min\{n_1, n_2\}$ (see
Table~\ref{tab:summary}). This is because: (1) using more cross pairs
did not substantially improve the efficiency curves in
Figure~\ref{fig:pareto}, (2) the CVXPY solver for binary-valued
problems would become unstable when using individual fairness if we
increase the number of cross pairs significantly. Furthermore, we used
the same cross validation folds and cross pairs when experimenting
with our different notions of fairness.

Finally, linear models perform poorly on some of our datasets~(see
e.g.~\cite{default-dataset}). However, our goal in this work was not
to obtain the most accurate models for particular datasets but to
study the trade-off between fairness and accuracy and also different
notions of fairness for regression problems.

\subsection{Datasets}
\label{sec:data-description}
The following is a more through description of the datasets used in this work.
We did not use any feature selection methods for any of the datasets. Although, for each dataset, 
we removed the uninformative features (e.g. various kinds of IDs). Categorical variables were converted 
to dummy/indicator variables, and an indicator for missing values were included per column with missing values.

\paragraph{The Adult dataset}
The Adult dataset~\citep{adult-dataset, uci-repo} from the UCI
repository contains Census data (hours worked per
week, education, sex, age, marital status, and so forth), and has been
used to train predictors as to whether an individual earns more or
less than \$50,000 per year. Predicting income based upon less
sensitive attributes, such as education and occupation, can be used to
aid in other economic decisions (related to lending or hiring). It is
therefore important to design models which are equally predictive of
income regardless of gender or race. We created groups for this dataset based
on gender. Furthermore, we extracted the data only from the Adult.data file.

\paragraph{The Communities and Crime dataset}
The Communities and Crime Dataset~\citep{communities-crime-dataset,
  uci-repo}, from the UCI repository is a dataset which includes many
features deemed relevant to violent crime rates (such as the
percentage of the community's population in an urban area, the
community's racial makeup, law enforcement involvement and racial
makeup of that law enforcement in a community, amount a community's
law enforcement allocated to drug units) for different
communities. This data is provided to train regression models based on
this data to predict the amount of violent crime (murder, rape,
robbery, and assault) in a given community. If police departments
allocate their units to communities based upon a prediction of this
form, this should draw concerns of fairness if the predictive behavior
is much better for violent predominantly white community than for a
violent predominantly nonwhite communities.  We consider communities
racial makeup as sensitive variable, however, the racial makeup
consists of more than two races.  We created two groups for this
dataset based on whether the percentage of black people in a community
(blackPerCap) is higher than the percentage of whites (whitePerCap),
indians (indianPerCap), asians (AsianPerCap) and hispanics
(HispPerCap) in that community.


\paragraph{The COMPAS dataset}
The COMPAS dataset contains data from Broward County, Florida originally compiled
by ProPublica~\cite{propublica} in which the goal
is to predict whether a convicted individual would commit a violent crime in the following two years
or not.
Following the analysis of Propublica and~\citet{Corbett-DaviesP17} we considered black
and white defendants who were assigned COMPAS risk scores
within 30 days of their arrest. Furthermore, we restricted ourselves to defendants who
``spent at least two years outside
a correctional facility without being arrested for a violent crime,
or were arrested for a violent crime within this two-year period~\cite{Corbett-DaviesP17}''.
Our goal is to predict the two-year violent recidivism. 
The dataset includes race, age, sex, number of prior convictions, and COMPAS violent crime
risk score (a score between 1 and 10 which we categorized to low, medium and high levels of risks).
As mentioned earlier, we created groups for this dataset based on race.

\paragraph{The Default dataset}
The Default of Credit Card Clients dataset~\citep{default-dataset, uci-repo} from the UCI
repository, contains data from Taiwanese credit card
users, such as their credit limit, gender, education, marital status,
history of payment, bill and payment amounts. These features are
designed to be used in the prediction of the probability of default
payments, although the data has binary labels. It has been
noted~\citep{default-dataset} that linear models tend to perform quite
poorly on this task. Given the sensitive nature of lending (and
historical inequity of the allocation of credit across races and
genders), it is important to understand the tradeoff between fairness
and accuracy in settings related to credit. We created groups for this 
dataset based on gender.

\paragraph{The Law School dataset}
The bar passage study was initiated in 1991 by Law School Admission Council national longitudinal. 
The dataset contains  records for law students who took the bar exam. 
The binary outcome indicates whether the student passed the bar exam or not. The features include variables such 
as cluster, lsat score, undergraduate GPA, zfyGPA, zGPA, full-time status, family income, age and also sensitive 
variables such as race and gender. 
The variable cluster is the result of a clustering of similar law schools (which is done apriori), and is used to adjust for the 
effect of type of law school. zGPA is the z-scores of the students overall GPA and zfyGPA is the  first year GPA relative to 
students at the same law school. We created groups for this dataset based on gender.
For more information about this dataset refer to \url{http://www2.law.ucla.edu/sander/Systemic/Data.htm}.

\paragraph{The Sentencing dataset}
The Sentencing dataset contains 5969 random sample of all prison inmates primarily from the department of corrections of a state. 
Moreover data from state agencies that have information of arrests are
also added to the dataset. To be sentenced to a state prison, an offender must have been convicted of a felony or must have pled guild to a felony. 
The outcome variable, sentence length, is the sentence given by a judge, but is not necessarily the amount of time served in prison. 
There are usually provisions for review by the state parole board, which can lead to release well before the full nominal sentence is served. 
The predictors include variables that are legitimate factors in sentencing (e.g., the crime for which a conviction was obtained) as well as sensitive variables 
like gender.  We created groups for this dataset based on gender.

\end{document}